\documentclass[runningheads]{llncs}

\usepackage[camera-ready]{eccv}

\usepackage{eccvabbrv}
\usepackage{wrapfig}
\usepackage{array}
\usepackage{graphicx}
\usepackage{booktabs}
\usepackage{multirow}
\usepackage{arydshln}
\usepackage{xcolor}
\usepackage{float}
\usepackage{enumitem}
\usepackage{boldline}
\usepackage{amsmath}
\usepackage[accsupp]{axessibility}  

\usepackage{hyperref}
\usepackage{hhline}
\usepackage{makecell}
\usepackage{orcidlink}

\begin{document}

\title{Q-REAL: Towards Naturalness and Distortion Evaluation for AI-Generated Content} 

\titlerunning{Q-REAL}

\author{Shushi Wang\inst{1} \and Zicheng Zhang\inst{3} \and
Chunyi Li\inst{1} \and Wei Wang\inst{2} \and Liya Ma\inst{2} \and Fengjiao Chen\inst{2} \and Xiaoyu Li\inst{2} \and Xuezhi Cao\inst{2} \and Guangtao Zhai\inst{1} \and Xiaohong Liu\inst{1,4}\textsuperscript{\textdagger}}

\authorrunning{Shushi Wang.~Author et al.}

\institute{Shanghai Jiao Tong University \and
Meituan \and Shanghai AI Lab \and Shanghai Innovation Institute}

\maketitle

\renewcommand{\thefootnote}{\textdagger}
\footnotetext{Corresponding author.}
\renewcommand{\thefootnote}{\arabic{footnote}}
\begin{abstract}
Quality assessment of AI-generated content is vital for model optimization, yet most existing evaluation datasets and models rely on coarse-grained single scores, failing to provide targeted guidance. To bridge this gap, we introduce \textbf{Q-Real}, a fine-grained quality assessment dataset specifically designed for AI-generated images. This dataset comprises 10K images generated by multiple models, annotated along two critical dimensions, \textbf{naturalness} and \textbf{distortion} which are widely regarded as the most significant aspects of AI-generated image quality. For each image, we localize major entities and provide a set of judgment questions and attribution descriptions along these dimensions to facilitate comprehensive evaluation. Based on this dataset, we establish the \textbf{Q-Real Bench} to rigorously evaluate models on the challenging tasks of judgment and grounding with reasoning. And to tackle these challenges, we further propose a fine-grained training pipeline for Multimodal Large Language Models (MLLMs), empowering them to judge, localize problematic entities with detailed analysis, and predict quality score. Experimental results demonstrate the high quality and significance of our dataset, as well as the effectiveness of our proposed training pipeline. Dataset and code will be released upon publication.
\end{abstract}

\section{Introduction}
\label{sec:intro}

With the rapid advancement of generative AI, massive amounts of text-to-image content are being produced and applied across various domains~\cite{everypixel2024, AImarket2024}. However, current generative content sometimes exhibits issues, which necessitate accurate quality assessment models to assess the generated content before practical use. Such evaluation models are also critical for guiding the iterative optimization of generative models. Existing studies have established a comprehensive evaluation paradigm for AI-generated images, encompassing text–image alignment, image quality, and visual aesthetics, with effective models available for each evaluation dimension~\cite{Kirstain2023pick, xu2023imagereward, li2023agiqa, zhang2025qeval, cao2025artimuse, jiang2024genai}. However, most existing evaluation approaches remain relatively coarse-grained, typically compressing complex generation results into a single score, which makes it difficult to precisely identify the sources of errors. Due to the lack of interpretable evaluation signals, such methods provide limited guidance for the targeted optimization of generative models. Consequently, developing fine-grained evaluation models is of critical importance.

\begin{figure*}[!t]
    \centering
    \centerline{\hspace*{-0.2cm}\includegraphics[scale=0.60]{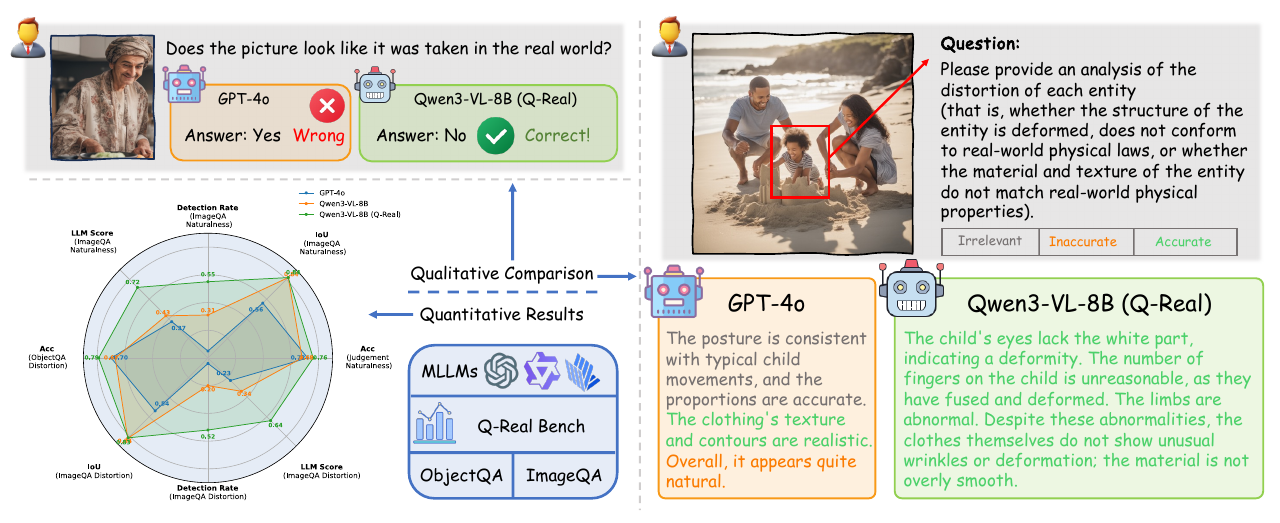}}
    \caption{Abilities of Q-Real-tuned Qwen3-VL-8B-Instruct~\cite{chen2024expanding} on fine-grained quality assessment tasks about AI-generated images, in comparison with the baseline version.}
    \vspace{-0.6cm}
    \label{fig:overall}
\end{figure*}

Given that mainstream AI-generated image applications predominantly target photorealism, our work focuses on constructing a fine-grained quality evaluation framework tailored for this scenario. Within the context of photorealistic generation, we evaluate image quality along two complementary dimensions: \textbf{naturalness}, which measures perceptible artificiality or AI-related artifacts (often referred to as the "AI look"), and \textbf{distortion}, which focuses on visual degradations such as blur, deformation, and structural corruption.

To support the construction of the proposed fine-grained quality evaluation framework, we construct a dataset of 10K AI-generated images with annotations on the two dimensions of naturalness and distortion, named \textbf{Q-Real}. Specifically, 7K images are sourced from the existing Q-Eval-100K dataset~\cite{zhang2025qeval}, while the remaining 3K are generated by more recent state-of-the-art models~\cite{flux11pro2025, Kolors2025, Hmrishav2025sd35, Dreamina2025}. For fine-grained annotation along these two dimensions, we localize major entities within each image using bounding boxes. For each entity, we design corresponding objective judgment questions for both naturalness and distortion. Furthermore, for entities identified as problematic, we provide detailed descriptive analyses of the specific issues. Given the high labor cost and time-consuming nature of such fine-grained annotation, drawing inspiration from prior works~\cite{chen2024groundingiqa, meng2025identity}, we design an automated annotation pipeline combined with manual refinement. This approach ensures both the efficiency of dataset construction and the high quality of the annotations.

Based on this dataset, we construct the benchmark \textbf{Q-Real Bench}, featuring two complementary tasks: \textbf{ObjectQA} and \textbf{ImageQA}. ObjectQA requires models to answer judgment questions about naturalness and distortion of entities in images, which assesses the fundamental ability to discriminate naturalness and distortion in AI-generated images, enabling efficient filtering of low-quality entities. In contrast, ImageQA requires models to localizing problematic entities and provide detailed description, which demands a comprehensive understanding to localize and explain these issues, providing detailed feedback crucial for optimizing generative models.

To tackle the challenges posed by Q-Real Bench, we turn to Multimodal Large Language Models (MLLMs)~\cite{bai2025Qwen25vl, chen2024expanding, bai2025Qwen3-VL, liu2024llavanext, gpt4o2024}, which have demonstrated remarkable capabilities in general visual tasks~\cite{zhang2023xcomposer}. However, their performance remains suboptimal in fine-grained quality assessment, particularly when discerning subtle naturalness issues and distortions, as shown in Figure~\ref{fig:overall}. To maximize evaluation capability in this domain,  we propose a \textbf{three-stage progressive training pipeline}. In the first stage, we fine-tune the model using object-level judgment QA pairs to establish a fundamental understanding of naturalness and distortion. Building upon this, the second stage leverages image-level grounding and reasoning descriptions to enable fine-grained evaluation. Finally, we employ the model's detailed assessments as Chain-of-Thought (CoT)~\cite{wei2022cot}reasoning context prompt, combining them with a rating token mapping strategy to predict quality scores, resulting in a comprehensive quality assessment model capable of simultaneous judgment, localization, analysis, and scoring. 

\noindent The main contributions of this paper are summarized as follows:

\begin{itemize}[leftmargin=*, label=\textbullet]
    \item We present \textbf{Q-Real}, the first fine-grained quality evaluation dataset for AI-generated images on \textbf{Naturalness} and \textbf{Distortion}, built via an automated pipeline with manual refinement.
    \item We design \textbf{Q-Real Bench}, comprising \textbf{ObjectQA} and \textbf{ImageQA}, to offer a comprehensive evaluation framework that systematically examines the capacity of MLLMs in fine-grained quality assessment.
    \item We propose a \textbf{three-stage training pipeline} and conduct experiments on multiple MLLMs, demonstrating the high quality of the dataset, the comprehensiveness of the benchmark, and the effectiveness of our approach in enhancing fine-grained evaluation.
\end{itemize}

\section{Related Work}

\subsection{Coarse-grained Image Quality Assessment}
Early image quality assessment (IQA) methods, including both full-reference and no-reference approaches, were primarily developed for user-generated images and aimed to predict overall perceptual quality by modeling human visual perception. Traditional metrics and early CNN-based models such as NIQE~\cite{mittal2013niqe} and DBCNN~\cite{zhang2020blind}, as well as transformer-based methods like LIQE~\cite{zhang2023liqe} and CLIP-IQA~\cite{wang2022clipiqa}, typically provide a single quality score. More recently, MLLM-based approaches such as Q-Align~\cite{wu2023qalign} and DEQA~\cite{you2024deqa_score} have been introduced for IQA, but most still follow a coarse-grained evaluation paradigm.

With the development of generative models like GANs~\cite{goodfellow2014generative}, diffusion models~\cite{ho2020denoising}, and autoregressive models~\cite{tian2024VAR}, the evaluation of generated images has increased. Early evaluation of AI-generated images mainly relied on objective metrics such as PSNR~\cite{gonzalez2002digital} and LPIPS~\cite{zhang2018unreasonable}, which focus on pixel-wise or perceptual feature differences. To better reflect human perception, several subjectively annotated datasets have been introduced, including AGIQA-3K~\cite{li2023agiqa}, AIGIQA-20K~\cite{li2024aigiqa}, and Q-Eval-100K~\cite{zhang2025qeval}, which provide mean opinion scores (MOS), as well as Pick-a-Pic~\cite{Kirstain2023pick} and HPS~\cite{wu2024human, wu2024humanv3}, which use side-by-side annotations. These datasets have enabled the adaptation of conventional IQA models for assessing AI-generated images. However, most existing methods and datasets remain coarse-grained,as they primarily predict a single overall quality score and lack the ability to provide analyses of quality issues in AI-generated images.
\vspace{-0.4cm}
\subsection{Fine-grained Image Quality Assessment}
To overcome the limitation that score-based evaluation fails to reveal the underlying causes of image quality degradation, fine-grained image quality assessment has been increasingly studied.

Among existing fine-grained image quality assessment efforts for user-generated content, Q-Instruct~\cite{wu2024qinstruct} represents a representative attempt to extend quality evaluation beyond scalar scores. Specifically, it formulates fine-grained image quality assessment as a set of quality-related questions, including objective multiple-choice and true–false tasks, as well as overall quality judgments with explanatory reasoning. It further incorporates quality-aware reasoning by posing questions derived from image quality evaluation, such as suggesting adjustments to shooting conditions to improve image quality. Grounding-IQA~\cite{chen2024groundingiqa} further enriches the expression of fine-grained quality assessment by introducing spatial grounding. Instead of only providing textual or scalar evaluations, it requires models to localize regions with quality issues by outputting bounding boxes, thereby explicitly indicating where visual degradations occur within an image.

For AI-generated images, RichHF-18K~\cite{liang2024rich} pioneered fine-grained quality assessment by linking quality with plausibility, providing scores and pixel-level heatmaps to localize issues. Despite this progress, existing methods still struggle to jointly capture description and spatial grounding. This limitation motivates our Q-Real, which integrates grounding with description-based evaluation.

\section{Q-Real Dataset Construction}
\subsection{Source Collection}

\noindent \textbf{\textit{Data Collection.}} To construct a high-quality dataset for our task, we collected images from two sources. First, we sampled a subset from Q-Eval-100K~\cite{zhang2025qeval}, a dataset featuring diverse images generated by multiple models across various categories and quality levels. Second, considering the continuous iteration of generative models, and to ensure the quality and timeliness of the dataset, we used the latest models including flux.1.1pro~\cite{flux11pro2025}, Kwai-Kolors~\cite{Kolors2025}, sd3.5~\cite{Hmrishav2025sd35} and dreamina3.0~\cite{Dreamina2025} to generate additional images.

\begin{figure*}[htbp]
    \centering
    \centerline{\hspace*{-0.2cm}\includegraphics[scale=0.75]{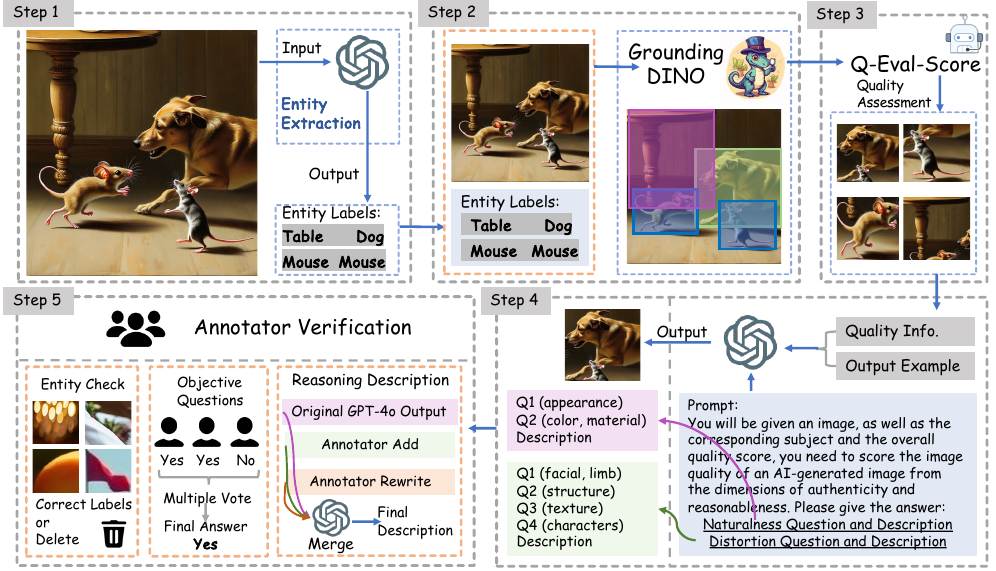}}
    \caption{The pipeline of our dataset construction.}
    \vspace{-0.6cm}
    \label{fig:dataset_pipeline}
\end{figure*}

\noindent \textbf{\textit{Prompt Selection.}} Whether sampling images from the Q-Eval-100K dataset or generating new ones using models, selecting appropriate prompts is crucial to ensuring the comprehensiveness and diversity of the dataset. When sampling from Q-Eval-100K, we designed four types of prompts (portrait generation, character-animal generation, entity generation, and scene generation) and filtered the images based on these prompts to cover a variety of generative tasks. For the generated portion, we created relevant prompts for these four categories using GPT, and then used them to produce images, thereby guaranteeing the dataset’s representativeness and diversity.

\subsection{Q-Real Annotation Pipeline}
A high-quality dataset is essential for training accurate evaluation models and establishing a reliable evaluation benchmark. However, annotating such a dataset with subjective description and bounding boxes is labor-intensive and costly. To address this challenge, we have developed a hybrid annotation strategy that combines automated annotation with manual verification and supplement, shown in Figure~\ref{fig:dataset_pipeline}. The process is as follows:

\noindent \textbf{\textit{Entity Name Extraction.}} Initially, we provide both the generated images and their corresponding prompts to GPT-4o~\cite{gpt4o2024} to extract the names of entities appearing in the images, ensuring comprehensive entity identification. By jointly analyzing visual content and textual context, GPT-4o can more accurately capture entities present in the generated images.

\noindent \textbf{\textit{Object Detection.}} For each entity, we employ Grounding DINO~\cite{liu2023grounding} for precise localization, which offers superior reliability compared to direct MLLM extraction. To ensure annotation quality, we filter out low-confidence detections and extremely small, visually ambiguous bounding boxes.

\noindent \textbf{\textit{Quality Evaluation.}}
Following detection, we use the AI-generated image quality assessment model Q-Eval-Score~\cite{zhang2025qeval} to assess the quality of each detected object. The quality score serves as an important conditioning signal for GPT-4o or other MLLMs to generate accurate analyses of naturalness and distortion. By incorporating quality scores, the model is guided to produce more reliable and well-aligned descriptions of the visual quality of detected entities.

\noindent \textbf{\textit{Generate Question-Answer Pairs and Reasoning Descriptions.}} Lastly, each extracted object, together with its corresponding quality score, is fed into GPT-4o to generate objective judgments and reasoning descriptions of its naturalness and distortion. Through this automated annotation pipeline comprising the first four steps, we obtain the initial version of the Q-Real dataset. 

\noindent \textbf{\textit{Subjective Experiment.}} After the automated annotation pipeline, a manual verification step is conducted to ensure annotation quality. 


First, a trained annotator verifies whether each extracted object matches its assigned label, correcting any mismatches. Objects with bounding boxes that are too small or visually unclear are removed.

After this initial verification, each retained object is independently reviewed by three annotators. They first determine whether the GPT-4o-generated description contains errors related to naturalness or distortion. If errors are identified, all annotators independently write human descriptions to replace the generated one; if not, they may supplement missing details. The final subjective description is synthesized by merging annotator inputs with an LLM, ensuring that multiple perspectives are considered and the description is as comprehensive as possible. For objects with erroneous descriptions, annotators also answer objective judgment questions, and the majority vote is used as the final label. This multi-step protocol ensures accurate fine-grained annotation.
\begin{wrapfigure}{r}{0.5\textwidth} 
    \vspace{-12pt} 
    \centering
    \includegraphics[width=0.48\textwidth]{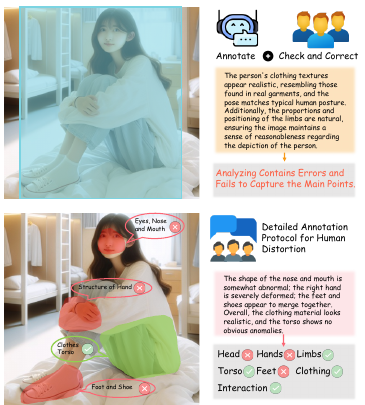}
    \caption{An example of finer-grained annotation for the distortion of human portrait.}
    \label{fig:human}
    \vspace{-20pt} 
\end{wrapfigure}

Additionally, considering that human portrait generation is comparatively more challenging and involves diverse application scenarios, we provide finer-grained evaluation annotations for the distortion dimension of 400 human-centric images, shown in Figure~\ref{fig:human}. Specifically, each human entity is analyzed across multiple body and semantic dimensions, including face, hands, limbs, torso, feet, clothing, and interactions with the surrounding environment. For each of these dimensions, three annotators independently examine the entity and mark the specific issues observed, ensuring both granularity and consistency of the evaluation.

\begin{figure*}[!t]
    \centering
    \centerline{\hspace*{0.1cm}\includegraphics[scale=0.73]{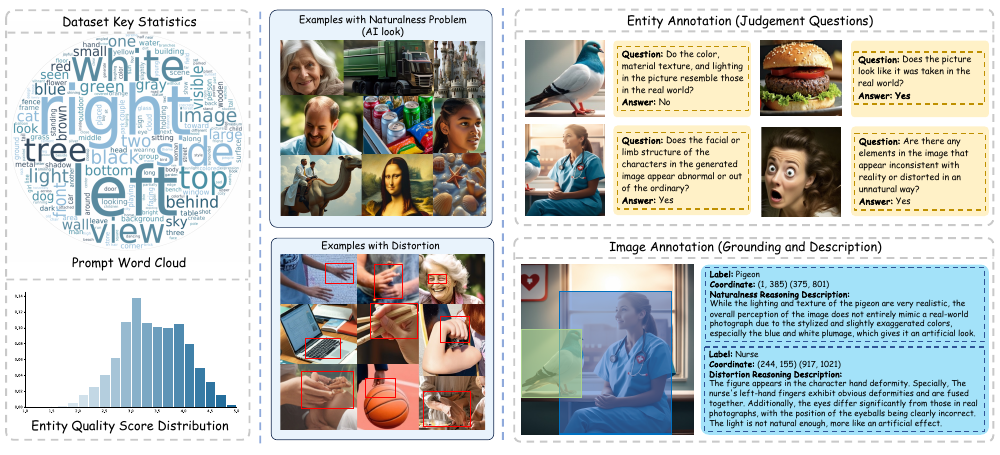}}
    \caption{Overview of the Q-Real dataset. Left: Word clouds and quality score distributions showcasing diversity. Middle: Representative examples of naturalness and distortion issues. Right: Sample images with annotations.}
    \vspace{-0.6cm}
    \label{fig:dataset}
\end{figure*}
\subsection{Dataset Format}
\label{sec:dataset_judgment}
Our dataset consists of 10K AI-generated images, with 7K sampled from the Q-Eval-100K dataset and 3K generated using the latest models. All images were generated using a total of 3,049 prompts. For each image, we marked all the main entities using bounding boxes. In total, we annotated 32,815 entities, covering various categories such as people, animals, and objects. Figure~\ref{fig:dataset} provides an overview of the Q-Real, illustrating its diversity and representative examples.

For all annotated entities, we provide fine-grained quality annotations under the two dimensions, \textbf{naturalness} and \textbf{distortion}. Each dimension includes both objective judgment questions and a subjective attribution description, enabling structured evaluation as well as explanatory analysis.

For each entity, two judgment questions are constructed to evaluate its naturalness:

\begin{enumerate}
\item \textit{Does the entity appear as if it exists in the real world?}
\item \textit{Do the color, material, texture, and lighting of the entity resemble those observed in the real world?}
\end{enumerate}

\noindent In addition, four judgment questions are formulated to assess the distortion of the entity:
\begin{enumerate}
\item \textit{Does the structural configuration (e.g., facial features or limbs) of the entity appear abnormal or anatomically implausible?}
\item \textit{Are there any unnatural deformations or inconsistencies in the shape of the entity?}
\item \textit{Does the material or texture of the entity deviate from realistic physical properties?}
\item \textit{If text is present within the entity, does it contain errors or illegible characters?}
\end{enumerate}

For the subjective attribution descriptions, the naturalness dimension reflects overall visual naturalness and whether the entity appears AI-generated ("AI look"), focusing on aspects such as color, lighting, and material realism. In contrast, the distortion dimension describes whether there are regions with structural abnormalities, blurriness, or physically implausible features. Compared to RichHF-18K~\cite{wu2024human}, which provides coarse heatmap annotations for distortion regions and lacks descriptive analysis, our dataset is more fine-grained and includes more accurate localization and detailed attribution descriptions. More detailed information about Q-Real can be found in the Supplementary Material. 

\begin{figure*}[!t]
    \centering
    \centerline{\hspace*{-0.2cm}\includegraphics[scale=0.78]{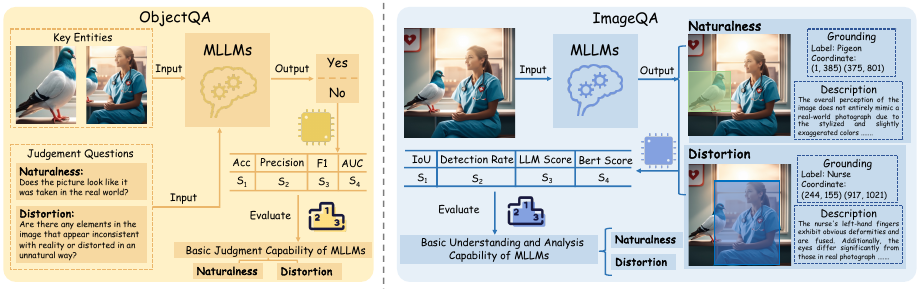}}
    \caption{Overview of Q-Real Bench, illustrating its two tasks: ObjectQA and ImageQA, including evaluation procedures and metrics.}
    \vspace{-0.6cm}
    \label{fig:benchmark}
\end{figure*}

\subsection{Q-Real Bench}
For evaluating model performance in model performance in fine-grained assessment of AI-generated images along the naturalness and distortion, we construct a benchmark, Q-Real Bench, as shown in Figure~\ref{fig:benchmark}. The Q-Real Bench includes 1,000 images of various types and quality, which are selected from Q-Real Dataset. Based on this benchmark, we evaluate the
performance of models from two perspectives: 

\noindent \textbf{\textit{ObjectQA.}} This task requires the model to answer the objective judgment questions annotated in Q-Real for each entity in an AI-generated image, evaluating its basic ability to assess naturalness and distortion at the entity level. Since ObjectQA involves only binary (Yes/No) responses, we adopt four standard metrics—Accuracy (\textit{Acc}), \textit{Precision}, \textit{F1}, and the Area Under the ROC Curve (\textit{AUC})—to comprehensively evaluate performance.

\noindent \textbf{\textit{ImageQA.}} The ImageQA task requires the model to localize problematic entity regions in the image under the naturalness and distortion dimensions, and provide corresponding descriptions of the issues. 

To evaluate the model's grounding performance, we adopt two standard metrics: Intersection over Union (\textit{IoU}) and \textit{Detection Rate}. For computing IoU, we use the Hungarian matching algorithm to establish an optimal one-to-one correspondence between predicted and ground-truth entities. Given a set of predicted bounding boxes $\mathcal{P} = \{p_i\}_{i=1}^{N}$ and ground-truth boxes $\mathcal{G} = \{g_j\}_{j=1}^{M}$, we construct a cost matrix $C \in \mathbb{R}^{N \times M}$ based solely on coordinate overlap, and solve for the optimal matching permutation $\pi^*$ as follows:
\begin{equation}
C_{ij} = 1 - \text{IoU}(p_i, g_j), \quad
\pi^* = \arg \min_{\pi} \sum_{i=1}^{K} C_{i\pi(i)}
\end{equation}
where $K = \min(N, M)$, and $C_{i\pi(i)}$ denotes the cost between the $i$-th predicted and matched ground-truth box. The final matched pairs are then used to compute the IoU.
Detection Rate is defined as the fraction of predicted entities that both matching a ground-truth box with IoU greater than 0.5 and providing a description that is consistent with the annotated issue (LLM Score is greater than 0.5). \textit{LLM Score} is introduced later.

To evaluate reasoning descriptions, we adopt two complementary metrics: \textit{LLM-Score} and \textit{Bert-Score}.
LLM-Score~\cite{Thakur2024llmjudge, zhang2024qbenchvideo} feeds the predicted and ground-truth descriptions into GPT-4o to measure semantic consistency, enabling robust evaluation beyond surface wording differences. We additionally report Bert-Score~\cite{zhang2020bertscore} as a traditional lexical overlap metric. Together, they provide complementary views of semantic correctness and textual similarity.

\section{Fine-grained Quality Assessment Training Pipeline}
\begin{figure*}[!t]
    \centering
    \centerline{\hspace*{-0.0cm}\includegraphics[scale=0.58]{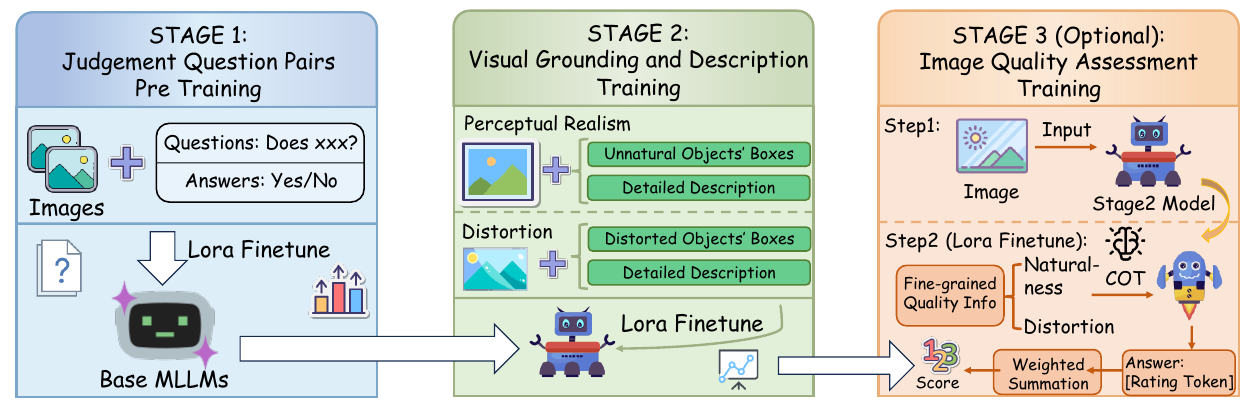}}
    \caption{Our proposed fine-grained quality assessment training pipeline.}
    \vspace{-0.6cm}
    \label{fig:model_design}
\end{figure*}

To empower MLLMs with comprehensive quality assessment capabilities, we propose a three-stage progressive training pipeline. This pipeline systematically enhances the model's ability to judge, localize, reason, and score AI-generated images, as illustrated in Figure~\ref{fig:model_design}.
\vspace{-0.4cm}
\subsection{Stage 1: Object-Level Judgment Training}
We fine-tune MLLMs using object-level judgment QA pairs from Q-Real to establish fundamental discriminative abilities for naturalness and distortion. For each key entity, we utilize six question-answer pairs, formatted as follows:

\noindent \textit{\textquotedblleft query\textquotedblright: \textquotedblleft \textless image\textgreater Does the entity appear as if it exists in the real world?\textquotedblright}

\noindent \textit{\textquotedblleft response\textquotedblright: \textquotedblleft Yes/No\textquotedblright}

\vspace{-0.4cm}
\subsection{Stage 2: Fine-Grained Evaluation Capability Training}
We construct a training dataset where each instance pairs a complete AI-generated image with the coordinates (bounding boxes) and detailed reasoning descriptions of problematic entities. To focus the model on discerning defects, we exclusively include entities exhibiting naturalness or distortion issues. This targeted approach enhances the model's capability to localize and analyze specific quality degradations. Taking the naturalness dimension as an example, the training instance is formatted as follows:

\noindent \textit{\textquotedblleft query\textquotedblright: \textquotedblleft \textless image\textgreater Analyze this AI-generated image and identify all entities that exhibit naturalness issues. For each such entity, provide its type label, its coordinates within the image and a brief explanation describing why the entity appears unrealistic.\textquotedblright}

\noindent \textit{\textquotedblleft response\textquotedblright: \textquotedblleft \textless label\textgreater\textless coordinate\textgreater\textless description\textgreater ... (repeated for each problematic entity)\textquotedblright}

\subsection{Stage 3: Quality Score Prediction Training}
\noindent \textit{\textbf{Context Prompt.}} Prior work on LMM-based quality assessment has demonstrated that employing a structured Chain-of-Thought (CoT) prompt can effectively guide models to yield more accurate quality evaluations~\cite{zhang2025qeval}. Inspired by this, we leverage the model fine-tuned in Stage 2 to first identify issues. This approach is grounded in two key observations: image quality is determined by the level of naturalness and the severity of distortion, and it is also heavily dependent on the quality of major entities within the image. Then, we incorporate the detailed fine-grained evaluation results about naturalness and distortion (\textless detected\_issues\textgreater) into the prompt as a structured COT.

\noindent \textit{\textless image\textgreater Please evaluate the overall quality of this image and assign it one of the following ratings: [Excellent, Good, Fair, Poor, Bad]. Reference Information: The following list details specific regions in the image identified as problematic. Each entry includes the bounding box coordinates of the defective area and a brief description of the issue: \textless detected\_issues\textgreater. Based on the visual content and the identified defects above, output only one of the ratings: [Excellent, Good, Fair, Poor, Bad]. Do not provide any reasoning.}

To ensure consistency between training and inference, we do not directly utilize ground-truth annotations to construct the prompt during this stage. Instead, we employ the model fine-tuned in Stage 2 to perform inference, dynamically generating localized defects and reasoning descriptions to form the prompt.

\noindent \textit{\textbf{Scoring Method.}} Following established practices~\cite{wu2023qalign, zhang2024qboost}, we map numerical MOS values to discrete adjective ratings for training, as LLMs interpret these more effectively. Specifically, we discretize the MOS range $[1, 5]$ into five intervals corresponding to $\{Bad, Poor, Fair, Good, Excellent\}$. The mapping function $R(s)$ is defined as:
\begin{equation}
    R(s) = r_i, \quad \text{if } m + \frac{i-1}{5}(M-m) < s \le m + \frac{i}{5}(M-m),
\end{equation}
where $\{r_i\}_{i=1}^5 = \{Bad, Poor, Fair, Good, Excellent\}$, $m=1$, and $M=5$.

During inference, we compute the continuous quality score $\hat{r}$ by extracting logits of the five rating tokens. We apply softmax to obtain probabilities $p_j$ and calculate the expectation:

\begin{equation}
    p_j = \frac{\exp(z_j)}{\sum_{k=1}^{5} \exp(z_k)}, \quad
    \hat{r} = \sum_{j=1}^{5} p_j \cdot w_j,
\end{equation}
where $w_j = \{0, 0.25, 0.5, 0.75, 1\}$ are weights for ratings from Bad to Excellent.

\noindent \textit{\textbf{Loss Function.}} To optimize the model for both general question-answering capabilities and precise score prediction, we adopt a hybrid loss function combining Cross-Entropy (CE) Loss and Mean Squared Error (MSE) Loss. The CE Loss, $\mathcal{L}_{CE}$, facilitates learning the QA format and necessary knowledge, while the MSE Loss, $\mathcal{L}_{MSE}$, refines the accuracy of the predicted quality score. The total loss $\mathcal{L}$ is defined as a weighted sum:
\begin{equation}
    \mathcal{L} = \alpha \mathcal{L}_{CE} + \beta \mathcal{L}_{MSE},
\end{equation}
where $\mathcal{L}_{CE} = -\sum_{i=1}^{N} y_i \log(p_i)$ represents the standard cross-entropy loss for answer tokens, and $\mathcal{L}_{MSE} = (\hat{r} - r_{MOS})^2$ measures the squared difference between the predicted score $\hat{r}$ and the ground-truth MOS label $r_{MOS}$. We set the weights $\alpha$ and $\beta$ to 1 by default.
\section{Experiments}
\subsection{Experiment Setup}
\noindent \textit{\textbf{Q-Real Bench Evaluation.}} We evaluate representative models across three categories: closed-source (GPT-4o)~\cite{gpt4o2024}, quality assessment-specific (Q-Eval-Score) \cite{zhang2025qeval}, and open-source MLLMs (including Qwen3-VL-8B~\cite{bai2025Qwen3-VL}, Qwen2.5-VL-7B~\cite{bai2025Qwen25vl}, InternVL3-8B~\cite{zhu2025internvl3}, and InternVL2.5-8B~\cite{chen2024expanding}).  For the open-source models (Qwen and InternVL series), we compare their performance in both zero-shot settings and after fine-tuning with our proposed Stage 1 and Stage 2 training.

\noindent \textit{\textbf{Quality Assessment Evaluation.}} We employ Qwen3-VL-8B as the backbone model, fine-tuned using our proposed three stages training pipeline. We evaluate its performance on both Q-Real Bench and the existing AGIQA-3K dataset~\cite{li2023agiqa}, using SRCC and PLCC as metrics and further test on ImageReward~\cite{xu2023imagereward} with accuracy as the metric.

\noindent \textit{\textbf{Training Settings.}} We fine-tune the models for 3, 5, and 3 epochs in Stages 1, 2, and 3, with batch sizes of 4, 2, and 4, respectively. All other hyperparameters follow the default settings of the base models. Experiments are conducted on 8 NVIDIA H100 (80GB) GPUs using PyTorch.

\begin{table*}[!htbp]
\caption{Performance comparison on the \textbf{ObjectQA} task. Evaluation metrics include Acc $\uparrow$, Precision $\uparrow$, F1 score $\uparrow$, and AUC $\uparrow$. The models marked with * indicate models fine-tuned using Q-Real. And questions refer to the judgement questions in the naturalness and distortion dimensions, as defined in Section ~\ref{sec:dataset_judgment}.}

\centering
\renewcommand\arraystretch{1.48}
\setlength{\tabcolsep}{0.8pt}
\resizebox{\linewidth}{!}{
\begin{tabular}{l|cccc|cccc|cccc}
\hlineB{3}
\multirow{2}{*}{\textbf{Model}}
& \multicolumn{4}{c|}{\textbf{Naturalness Question 1}} & \multicolumn{4}{c|}{\textbf{Naturalness Question 2}} & \multicolumn{4}{c}{\textbf{Distortion Question 1}} \\ \cline{2-13}
& \textit{Acc} & \textit{Precision} & \textit{F1 Score} & \textit{AUC}
& \textit{Acc} & \textit{Precision} & \textit{F1 Score} & \textit{AUC}
& \textit{Acc} & \textit{Precision} & \textit{F1 Score} & \textit{AUC}\\
\hline
GPT-4o & 0.752 & 0.682 & 0.739 & - & 0.706 & 0.602 & 0.735 & - & 0.693 & 0.623 & 0.632 & - \\ \cdashline{1-13}
Q-Eval-Score & 0.675 & 0.723 & 0.613 & 0.741 & 0.645 & 0.636 & 0.634 & 0.688 & 0.683 & 0.555 & 0.552 & 0.711 \\ \cdashline{1-13}
Qwen2.5-VL-7B-Instruct  & 0.605 & 0.769 & 0.388 & 0.776 & 0.630 & 0.734 & 0.497 & 0.750 & 0.652 & 0.714 & 0.082 & 0.711 \\ 
Qwen2.5-VL-7B-Instruct$^{*}$ & $\text{0.751}_{\textcolor{red}{+14.6\%}}$ & $\text{0.716}_{\textcolor{gray}{-5.3\%}}$ & $\text{0.757}_{\textcolor{red}{+36.9\%}}$ & $\text{0.828}_{\textcolor{red}{+5.2\%}}$ 
& $\text{0.743}_{\textcolor{red}{+11.3\%}}$ & $\text{0.700}_{\textcolor{gray}{-3.4\%}}$ & $\text{0.757}_{\textcolor{red}{+26.0\%}}$ & $\text{0.826}_{\textcolor{red}{+7.6\%}}$ & $\text{0.770}_{\textcolor{red}{+12.2\%}}$ & $\text{0.685}_{\textcolor{gray}{-2.9\%}}$ & $\text{0.669}_{\textcolor{red}{+57.3\%}}$ & $\text{0.828}_{\textcolor{red}{+11.7\%}}$ \\ \cdashline{1-13}
InternVL2.5-8B & 0.700 & 0.692 & 0.687 & 0.285 & 0.651 & 0.602 & 0.699 & 0.341 & 0.656 & 0.574 & 0.221 & 0.683 \\

InternVL2.5-8B$^{*}$ & $\text{0.762}_{\textcolor{red}{+6.2\%}}$ & $\text{0.740}_{\textcolor{red}{+4.8\%}}$ & $\text{0.758}_{\textcolor{red}{+7.1\%}}$ & $\text{0.807}_{\textcolor{red}{+52.2\%}}$ 
& $\text{0.752}_{\textcolor{red}{+10.1\%}}$ & $\text{0.730}_{\textcolor{red}{+12.8\%}}$ & $\text{0.753}_{\textcolor{red}{+5.4\%}}$ & $\text{0.814}_{\textcolor{red}{+47.3\%}}$ & $\text{0.767}_{\textcolor{red}{+11.1\%}}$ & $\text{0.691}_{\textcolor{red}{+11.7\%}}$ & $\text{0.658}_{\textcolor{red}{+43.7\%}}$ & $\text{0.842}_{\textcolor{red}{+15.9\%}}$ \\ \cdashline{1-13}

Qwen3-VL-8B-Instruct  & 0.702 & 0.694 & 0.686 & 0.776 & 0.661 & 0.651 & 0.652 & 0.724 & 0.562 & 0.283 & 0.195 & 0.398 \\

Qwen3-VL-8B-Instruct$^{*}$ & $\text{0.762}_{\textcolor{red}{+6.0\%}}$ & $\text{0.728}_{\textcolor{red}{3.4\%}}$ & $\text{0.765}_{\textcolor{red}{+7.9\%}}$ & $\text{0.843}_{\textcolor{red}{+6.7\%}}$ 
& $\text{0.755}_{\textcolor{red}{+9.4\%}}$ & $\text{0.724}_{\textcolor{red}{+7.3\%}}$ & $\text{0.761}_{\textcolor{red}{+10.9\%}}$ & $\text{0.840}_{\textcolor{red}{+11.6\%}}$ & $\text{0.777}_{\textcolor{red}{+21.5\%}}$ & $\text{0.718}_{\textcolor{red}{+43.5\%}}$ & $\text{0.663}_{\textcolor{red}{+46.8\%}}$ & $\text{0.843}_{\textcolor{red}{+44.5\%}}$   \\ \cdashline{1-13}
InternVL3-8B & 0.682 & 0.690 & 0.653 & 0.776 & 0.643 & 0.591 & 0.702 & 0.748 & 0.659 & 0.596 & 0.227 & 0.714 \\

InternVL3-8B$^{*}$ & $\text{0.737}_{\textcolor{red}{+5.5\%}}$ & $\text{0.692}_{\textcolor{red}{0.2\%}}$ & $\text{0.655}_{\textcolor{red}{+0.2\%}}$ & $\text{0.810}_{\textcolor{red}{+3.4\%}}$ 
& $\text{0.734}_{\textcolor{red}{+9.1\%}}$ & $\text{0.700}_{\textcolor{red}{+10.9\%}}$ & $\text{0.646}_{\textcolor{red}{+6.4\%}}$ & $\text{0.805}_{\textcolor{red}{+5.7\%}}$ & $\text{0.794}_{\textcolor{red}{+13.5\%}}$ & $\text{0.752}_{\textcolor{red}{+15.6\%}}$ & $\text{0.767}_{\textcolor{red}{+54.0\%}}$ & $\text{0.874}_{\textcolor{red}{+16.0\%}}$   \\

\midrule \midrule

\multirow{2}{*}{}
& \multicolumn{4}{c|}{\textbf{Distortion Question 2}} & \multicolumn{4}{c|}{\textbf{Distortion Question 3}} & \multicolumn{4}{c}{\textbf{Distortion Question 4}} \\ \cline{2-13}
& \textit{Acc} & \textit{Precision} & \textit{F1 Score} & \textit{AUC}
& \textit{Acc} & \textit{Precision} & \textit{F1 Score} & \textit{AUC}
& \textit{Acc} & \textit{Precision} & \textit{F1 Score} & \textit{AUC}\\
\hline
GPT-4o & 0.603 & 0.591 & 0.632 & - & 0.605 & 0.728 & 0.441 & - & 0.929 & 0.700 & 0.688 & - \\ \cdashline{1-13}
Q-Eval-Score & 0.609 & 0.584 & 0.706 & 0.666 & 0.589 & 0.558 & 0.698 & 0.692 & 0.852 & 0.406 & 0.520 & 0.873 \\ \cdashline{1-13}
Qwen2.5-VL-7B-Instruct & 0.563 & 0.810 & 0.351 & 0.690 & 0.673 & 0.828 & 0.587 & 0.754 & 0.892 & 0.736 & 0.074 & 0.881 \\
Qwen2.5-VL-7B-Instruct$^{*}$ & $\text{0.702}_{\textcolor{red}{+23.9\%}}$ & $\text{0.719}_{\textcolor{gray}{-9.1\%}}$ & $\text{0.717}_{\textcolor{red}{+36.6\%}}$ & $\text{0.772}_{\textcolor{red}{+3.2\%}}$ 
& $\text{0.735}_{\textcolor{red}{+6.2\%}}$ & $\text{0.744}_{\textcolor{gray}{-8.4\%}}$ & $\text{0.738}_{\textcolor{red}{+15.1\%}}$ & $\text{0.812}_{\textcolor{red}{+5.8\%}}$ & $\text{0.933}_{\textcolor{red}{+4.1\%}}$ & $\text{0.689}_{\textcolor{gray}{-4.7\%}}$ & $\text{0.701}_{\textcolor{red}{+62.7\%}}$ & $\text{0.929}_{\textcolor{red}{+4.8\%}}$ \\ \cdashline{1-13}
InternVL2.5-8B & 0.608 & 0.604 & 0.668 & 0.554 & 0.554 & 0.536 & 0.680 & 0.402 & 0.896 & 0.648 & 0.223 & 0.886 \\
InternVL2.5-8B$^{*}$ & $\text{0.707}_{\textcolor{red}{+9.9\%}}$ & $\text{0.707}_{\textcolor{red}{+10.3\%}}$ & $\text{0.731}_{\textcolor{red}{+6.3\%}}$ & $\text{0.760}_{\textcolor{red}{+20.6\%}}$ 
& $\text{0.744}_{\textcolor{red}{+19.0\%}}$ & $\text{0.744}_{\textcolor{red}{20.8\%}}$ & $\text{0.752}_{\textcolor{red}{+7.2\%}}$ & $\text{0.819}_{\textcolor{red}{+41.7\%}}$ & $\text{0.945}_{\textcolor{red}{+4.9\%}}$ & $\text{0.798}_{\textcolor{red}{+15.0\%}}$ & $\text{0.727}_{\textcolor{red}{+50.4\%}}$ & $\text{0.935}_{\textcolor{red}{+4.9\%}}$ \\ \cdashline{1-13}

Qwen3-VL-7B-Instruct & 0.600 & 0.590 & 0.676 & 0.665 & 0.622 & 0.642 & 0.612 & 0.667 & 0.928 & 0.874 & 0.559 & 0.915 \\
Qwen3-VL-7B-Instruct$^{*}$ & $\text{0.709}_{\textcolor{red}{+10.9\%}}$ & $\text{0.719}_{\textcolor{red}{+12.9\%}}$ & $\text{0.727}_{\textcolor{red}{+5.1\%}}$ & $\text{0.787}_{\textcolor{red}{+12.2\%}}$ 
& $\text{0.736}_{\textcolor{red}{+11.4\%}}$ & $\text{0.747}_{\textcolor{red}{+10.5\%}}$ & $\text{0.739}_{\textcolor{red}{+12.7\%}}$ & $\text{0.821}_{\textcolor{red}{+11.4\%}}$ & $\text{0.946}_{\textcolor{red}{+1.8\%}}$ & $\text{0.801}_{\textcolor{gray}{-7.3\%}}$ & $\text{0.737}_{\textcolor{red}{+17.8\%}}$ & $\text{0.953}_{\textcolor{red}{+3.8\%}}$ \\   \cdashline{1-13}

InternVL3-8B & 0.589 & 0.689 & 0.507 & 0.677 & 0.624 & 0.718 & 0.434 & 0.748 & 0.898 & 0.804 & 0.185 & 0.933 \\
InternVL3-8B$^{*}$ & $\text{0.701}_{\textcolor{red}{+11.2\%}}$ & $\text{0.706}_{\textcolor{red}{+1.7\%}}$ & $\text{0.723}_{\textcolor{red}{+21.6\%}}$ & $\text{0.784}_{\textcolor{red}{+10.7\%}}$ 
& $\text{0.727}_{\textcolor{red}{+10.3\%}}$ & $\text{0.724}_{\textcolor{red}{+0.6\%}}$ & $\text{0.737}_{\textcolor{red}{+30.3\%}}$ & $\text{0.822}_{\textcolor{red}{+7.4\%}}$ & $\text{0.939}_{\textcolor{red}{+4.1\%}}$ & $\text{0.729}_{\textcolor{gray}{-7.5\%}}$ & $\text{0.715}_{\textcolor{red}{+53.0\%}}$ & $\text{0.958}_{\textcolor{red}{+2.5\%}}$ \\ 

\hlineB{3}
\end{tabular}
}
\label{tab:judgement_findings}
\end{table*}

\begin{table*}[!t]\small    
    \centering
    
    \renewcommand\arraystretch{1.58}
    \setlength{\tabcolsep}{10pt}
    \caption{Performance comparison of different models on the \textbf{ImageQA} task. Evaluation metrics include IoU $\uparrow$, Detection Rate$\uparrow$, LLM Score$\uparrow$, and Bert Score$\uparrow$. The models marked with * indicate fine-tuned models.}

    \vspace{-6pt}
   \resizebox{\linewidth}{!}{\begin{tabular}{l|cccc|cccc}
    \hlineB{3}
   \multirow{2}{40pt}{\textbf{Model}}  & \multicolumn{4}{c|}{\textbf{Naturalness}} & \multicolumn{4}{c}{\textbf{Distortion}}  \\ \cline{2-9}
  &\multicolumn{1}{c}{\textit{IoU}}&\multicolumn{1}{c}{\textit{Detection Rate}}&\multicolumn{1}{c}{\textit{LLM Score}}&\multicolumn{1}{c|}{\textit{Bert Score}}&\multicolumn{1}{c}{\textit{IoU}}&\multicolumn{1}{c}{\textit{Detection Rate}}&\multicolumn{1}{c}{\textit{LLM Score}}&\multicolumn{1}{c}{\textit{Bert Score}} \\ \hline
  GPT-4o & 0.566 & 0.051 & 0.376 & 0.334  & 0.548 & 0.041 & 0.231 & 0.261  \\ \cdashline{1-9}
  Q-Eval-Score & 0.835 & 0.165 & 0.353 & 0.208  & 0.838 & 0.038 & 0.249 & 0.328  \\ \cdashline{1-9}
  Qwen2.5-VL-7B-Instruct  & 0.875 & 0.271 & 0.409 & 0.344 & 0.829 & 0.059  & 0.230 & 0.387 \\ 
  Qwen2.5-VL-7B-Instruct$^{*}$ 
  & $\text{0.875}_{\textcolor{red}{+0.0\%}}$ & $\text{0.596}_{\textcolor{red}{+32.5\%}}$ & $\text{0.713}_{\textcolor{red}{+30.4\%}}$ & $\text{0.482}_{\textcolor{red}{+13.8\%}}$ & $\text{0.878}_{\textcolor{red}{+4.9\%}}$ & $\text{0.548}_{\textcolor{red}{+48.9\%}}$ & $\text{0.633}_{\textcolor{red}{+40.3\%}}$ & $\text{0.394}_{\textcolor{red}{+0.7\%}}$  \\ \cdashline{1-9}
  InternVL2.5-8B & 0.730 & 0.095 & 0.390 & 0.362 & 0.763 & 0.031 & 0.255 & 0.278  \\
  
  InternVL2.5-8B$^{*}$
  & $\text{0.825}_{\textcolor{red}{+9.5\%}}$ & $\text{0.491}_{\textcolor{red}{+39.6\%}}$ & $\text{0.675}_{\textcolor{red}{+28.5\%}}$ & $\text{0.441}_{\textcolor{red}{+7.9\%}}$ & $\text{0.817}_{\textcolor{red}{+5.4\%}}$ & $\text{0.448}_{\textcolor{red}{+41.7\%}}$ & $\text{0.593}_{\textcolor{red}{+34.8\%}}$ & $\text{0.366}_{\textcolor{red}{+8.8\%}}$  \\
  \cdashline{1-9}
  Qwen3-VL-8B-Instruct & 0.848 & 0.313 & 0.435 & 0.322 & 0.848 & 0.206 & 0.341 & 0.263  \\
  
  Qwen3-VL-8B-Instruct$^{*}$ 
  & $\text{0.834}_{\textcolor{gray}{-1.4\%}}$ & $\text{0.548}_{\textcolor{red}{+23.5\%}}$ & $\text{0.722}_{\textcolor{red}{+28.7\%}}$ & $\text{0.483}_{\textcolor{red}{+16.1\%}}$ & $\text{0.832}_{\textcolor{gray}{-1.6\%}}$ & $\text{0.523}_{\textcolor{red}{+31.7\%}}$ & $\text{0.639}_{\textcolor{red}{+29.8\%}}$ & $\text{0.385}_{\textcolor{red}{+12.2\%}}$  \\
  \cdashline{1-9}
  InternVL3-8B & 0.888 & 0.060 & 0.230 & 0.354 & 0.878 & 0.274 & 0.413 & 0.274  \\
  
  InternVL3-8B$^{*}$ 
  & $\text{0.829}_{\textcolor{gray}{-5.9\%}}$ & $\text{0.527}_{\textcolor{red}{+46.7\%}}$ & $\text{0.737}_{\textcolor{red}{+50.7\%}}$ & $\text{0.499}_{\textcolor{red}{+14.5\%}}$ & $\text{0.825}_{\textcolor{gray}{-5.3\%}}$ & $\text{0.481}_{\textcolor{red}{+20.7\%}}$ & $\text{0.645}_{\textcolor{red}{+23.2\%}}$ & $\text{0.391}_{\textcolor{red}{+11.7\%}}$  \\
   
    \hlineB{3}
\end{tabular}}
\vspace{-12pt}
    \label{tab:ImageQA_Findings}
\end{table*}

\subsection{Discussion \& General Findings}
\noindent \textit{\textbf{1)For Q-Real Bench Evaluation.}} The general performance on the ObjectQA and ImageQA is exhibited in Table~\ref{tab:judgement_findings} and Table~\ref{tab:ImageQA_Findings}. In the ObjectQA task, existing MLLMs demonstrate a certain capability in judging naturalness and distortion in AI-generated images, generally performing better on naturalness than on distortion. However, their performance in the distortion dimension is notably weak, particularly in identifying implausible regions and illegible characters, where they show limited discriminative ability (characterized by high precision but low recall). After fine-tuning with Q-Real, the models exhibit a significant improvement in both dimensions. Most evaluation metrics across various questions exceed 0.7, substantially outperforming the zero-shot performance of MLLMs and even surpassing closed-source models like GPT-4o.

In the ImageQA task, existing MLLMs perform poorly, often indiscriminately detecting all entities in the image without accurately determining whether they exhibit naturalness or distortion issues. As a result, while IoU scores are high, both detection rate and description similarity remain low. After fine-tuning, models retain object detection ability and gains naturalness and distortion analysis, enabling accurate localization and detailed issue descriptions. Both detection rate and description similarity are significantly improved. Specifically, for the naturalness dimension, the detection rate reaches above 0.5 and the description similarity exceeds 0.7 (LLM Score); for the distortion dimension, the detection rate increases to 0.5 and the description similarity reaches 0.6 (LLM Score). Detailed visualization results are provided in the Supplementary Material.

\begin{wraptable}{r}{0.55\textwidth} 
    \vspace{-32pt} 
    \small
    \centering
    
    \renewcommand\arraystretch{1.28}
    \setlength{\tabcolsep}{6pt} 
    \caption{Performance comparison on quality assessment datasets. All models are trained on the subset of Q-Real derived from Q-Eval-100K. The best results are highlighted in \textbf{bold}, and the second-best are \underline{underlined}.}

    \resizebox{\linewidth}{!}{
        \begin{tabular}{l|cc|cc|c}
        \hlineB{2}
        \multirow{2}{*}{\textbf{Model}}  & \multicolumn{2}{c|}{\textbf{Q-Real}} & \multicolumn{2}{c|}{\textbf{AGIQA-3K}} & \multicolumn{1}{c}{\textbf{ImageReward}} \\ \cline{2-6}
        &\multicolumn{1}{c}{\textit{SRCC}}&\multicolumn{1}{c|}{\textit{PLCC}}&\multicolumn{1}{c}{\textit{SRCC}}&\multicolumn{1}{c|}{\textit{PLCC}}&\multicolumn{1}{c}{\textit{Acc}} \\ \hline
        CLIP-IQA~\cite{wang2022clipiqa}  & 0.264 & 0.262 & 0.074 & 0.088 & 0.493\\ 
        IPCE~\cite{Peng_2024_CVPR}  & 0.631 & 0.636 & 0.074 & 0.057 & 0.495\\ 
        Q-Align~\cite{wu2023qalign}  & 0.651 & 0.644 & 0.572 & 0.558 & 0.556\\ 
        Q-Eval-Score~\cite{zhang2025qeval}  & \underline{0.690} & \underline{0.693} & \underline{0.587} & \underline{0.582} & \underline{0.601}\\
        Q-Real-Score  & \textbf{0.710} & \textbf{0.720} & \textbf{0.608} & \textbf{0.611} & \textbf{0.607}\\
        \hlineB{2}
        \end{tabular}
    }
    \vspace{-15pt} 
    \label{tab:quality_assessment}
\end{wraptable}

\noindent \textit{\textbf{2)For Quality Assessment Evaluation.}} The general performance is exhibited in Table~\ref{tab:quality_assessment}.  Our model trained with the Stage 3, named Q-Real-Score, achieves the best performance on all Q-Real Bench, AGIQA-3K, and ImageReward. Compared to the second-best competitor (Q-Eval-Score), while it employs a structured CoT prompt to guide the model on which dimensions to consider for quality assessment, it only provides general instructions without details. In contrast, our approach incorporates detailed issues about naturalness and distortion into the prompt, offering more explicit guidance thereby improving performance.

\begin{wraptable}{r}{0.55\textwidth} 
    \vspace{-32pt} 
    \small
    \centering
    
    \renewcommand\arraystretch{1.44}
    \caption{Cross-dataset validation on RichHF-18K. The models marked with * indicate fine-tuned models.}

    \resizebox{\linewidth}{!}{
        \begin{tabular}{l|cc|cc}
        \hlineB{3}
        \multirow{2}{*}{\textbf{Model}}  & \multicolumn{2}{c|}{\textbf{ImageQA}} & \multicolumn{2}{c}{\textbf{ObjectQA}}  \\ \cline{2-5}
        &\multicolumn{1}{c}{\textit{Recall}}&\multicolumn{1}{c|}{\textit{Precision}}&\multicolumn{1}{c}{\textit{Acc}}&\multicolumn{1}{c}{\textit{F1 Score}} \\ \hline
        Qwen2.5-VL-7B-Instruct  & 0.102 & 0.122 & 0.493 & 0.492 \\ 
        Qwen2.5-VL-7B-Instruct$^{*}$  & $\text{0.401}_{\textcolor{red}{+29.9\%}}$ & $\text{0.531}_{\textcolor{red}{+40.9\%}}$ & $\text{0.551}_{\textcolor{red}{+5.8\%}}$ & $\text{0.518}_{\textcolor{red}{+2.6\%}}$ \\ \cdashline{1-5}
        Intern2.5VL-8B  & 0.221 & 0.175 & 0.464 & 0.441 \\ 
        Intern2.5VL-8B$^{*}$  & $\text{0.280}_{\textcolor{red}{+5.9\%}}$ & $\text{0.285}_{\textcolor{red}{+11.0\%}}$ & $\text{0.564}_{\textcolor{red}{+10.0\%}}$ & $\text{0.500}_{\textcolor{red}{+5.9\%}}$ \\ \cdashline{1-5}
        Qwen3-VL-8B-Instruct  & 0.505 & 0.273 & 0.532 & 0.404 \\ 
        Qwen3-VL-8B-Instruct$^{*}$  & $\text{0.537}_{\textcolor{red}{+3.2\%}}$ & $\text{0.615}_{\textcolor{red}{+34.2\%}}$ & $\text{0.556}_{\textcolor{red}{+2.4\%}}$ & $\text{0.492}_{\textcolor{red}{+8.8\%}}$ \\ \cdashline{1-5}
        Intern3VL-8B  & 0.172 & 0.175 & 0.442 & 0.465 \\ 
        Intern3VL-8B$^{*}$  & $\text{0.542}_{\textcolor{red}{+37.0\%}}$ & $\text{0.635}_{\textcolor{red}{+46.0\%}}$ & $\text{0.552}_{\textcolor{red}{+11.0\%}}$ & $\text{0.516}_{\textcolor{red}{+5.1\%}}$ \\ 
        \hlineB{3}
        \end{tabular}
    }
    \vspace{-20pt}
    \label{tab:cross_validation}
\end{wraptable}

\noindent \textit{\textbf{3)Cross-dataset Validation.}} In the RichHF-18K~\cite{liang2024rich}, the plausibility dimension is defined similarly to the distortion dimension in our work. To demonstrate the value of Q-Real, we conduct cross-dataset validation on the test data of RichHF-18K in the distortion dimension, with performance results shown in Tabel~\ref{tab:cross_validation}.

For ImageQA, we compute recall and precision between the annotated regions in RichHF-18K and the predicted boxes from our model. For ObjectQA, since RichHF-18K provides plausibility scores for each image, we consider images with scores above 0.6 as having no plausibility issues. We then use our model’s judgements on the distortion dimension and calculate the accuracy and F1 score between the two. Experimental results show that the model fine-tuned on Q-Real achieves accurate and stable performance on RichHF-18K, with clear improvements over the zero-shot baseline.

\begin{table*}[!t]\small    
    \centering
    
    \renewcommand\arraystretch{1.58}
    \setlength{\tabcolsep}{10pt}
    \caption{Ablation study on Stage 1 training. For each model, the $^1$ denotes training without Stage 1, while $^2$ indicates training with Stage 1 followed by Stage 2.}

    \vspace{-6pt}
   \resizebox{\linewidth}{!}{\begin{tabular}{l|cccc|cccc}
    \hlineB{3}
   \multirow{2}{40pt}{\textbf{Model}}  & \multicolumn{4}{c|}{\textbf{Naturalness}} & \multicolumn{4}{c}{\textbf{Distortion}}  \\ \cline{2-9}
  &\multicolumn{1}{c}{\textit{IoU}}&\multicolumn{1}{c}{\textit{Detection Rate}}&\multicolumn{1}{c}{\textit{LLM Score}}&\multicolumn{1}{c|}{\textit{Bert Score}}&\multicolumn{1}{c}{\textit{IoU}}&\multicolumn{1}{c}{\textit{Detection Rate}}&\multicolumn{1}{c}{\textit{LLM Score}}&\multicolumn{1}{c}{\textit{Bert Score}} \\ \hline
  Qwen2.5-VL-7B-Instruct$^{1}$  & 0.874 & 0.595 & 0.715 & 0.481 & 0.879 & 0.547  & 0.623 & 0.388 \\ 
  Qwen2.5-VL-7B-Instruct$^{2}$ 
  & $\text{0.875}_{\textcolor{red}{+0.1\%}}$ & $\text{0.596}_{\textcolor{red}{+0.1\%}}$ & $\text{0.723}_{\textcolor{red}{+0.8\%}}$ & $\text{0.482}_{\textcolor{red}{+0.1\%}}$ & $\text{0.878}_{\textcolor{gray}{-0.1\%}}$ & $\text{0.548}_{\textcolor{red}{+0.1\%}}$ & $\text{0.633}_{\textcolor{red}{+1.0\%}}$ & $\text{0.394}_{\textcolor{red}{+0.6\%}}$  \\ \cdashline{1-9}
  InternVL2.5-8B$^{1}$ & 0.820 & 0.482 & 0.667 & 0.435 & 0.808 & 0.418 & 0.577 & 0.356  \\
  
  InternVL2.5-8B$^{2}$
  & $\text{0.825}_{\textcolor{red}{+0.5\%}}$ & $\text{0.491}_{\textcolor{red}{+0.9\%}}$ & $\text{0.675}_{\textcolor{red}{+0.8\%}}$ & $\text{0.441}_{\textcolor{red}{+0.6\%}}$ & $\text{0.817}_{\textcolor{red}{+0.9\%}}$ & $\text{0.448}_{\textcolor{red}{+3.0\%}}$ & $\text{0.593}_{\textcolor{red}{+1.6\%}}$ & $\text{0.366}_{\textcolor{red}{+1.0\%}}$  \\
  \cdashline{1-9}
  Qwen3-VL-8B-Instruct$^{1}$ & 0.832 & 0.540 & 0.721 & 0.479 & 0.831 & 0.518 & 0.627 & 0.385  \\
  
  Qwen3-VL-8B-Instruct$^{2}$ 
  & $\text{0.834}_{\textcolor{red}{+0.2\%}}$ & $\text{0.548}_{\textcolor{red}{+0.8\%}}$ & $\text{0.722}_{\textcolor{red}{+0.1\%}}$ & $\text{0.483}_{\textcolor{red}{+0.4\%}}$ & $\text{0.832}_{\textcolor{red}{+0.1\%}}$ & $\text{0.523}_{\textcolor{red}{+0.5\%}}$ & $\text{0.639}_{\textcolor{red}{+1.2\%}}$ & $\text{0.385}_{\textcolor{red}{+0.0\%}}$  \\
  \cdashline{1-9}
  InternVL3-8B$^{1}$ & 0.830 & 0.522 & 0.728 & 0.488 & 0.821 & 0.477 & 0.629 & 0.385  \\
  
  InternVL3-8B$^{2}$ 
  & $\text{0.829}_{\textcolor{gray}{-0.1\%}}$ & $\text{0.527}_{\textcolor{red}{+0.5\%}}$ & $\text{0.737}_{\textcolor{red}{+0.9\%}}$ & $\text{0.499}_{\textcolor{red}{+1.1\%}}$ & $\text{0.825}_{\textcolor{red}{+0.4\%}}$ & $\text{0.481}_{\textcolor{red}{+0.4\%}}$ & $\text{0.645}_{\textcolor{red}{+1.6\%}}$ & $\text{0.391}_{\textcolor{red}{+0.6\%}}$  \\
   
    \hlineB{3}
\end{tabular}}
\vspace{-8pt}
    \label{tab:ImageQA_ablation}
\end{table*}

\begin{table}[!t]\small    
    \centering
    
    \renewcommand\arraystretch{1.28} 
    \setlength{\tabcolsep}{8pt}     
    \caption{Ablation study of different prompt on quality assessment. \textbf{N-CoT}: Naturalness CoT, \textbf{D-CoT}: Distortion CoT, \textbf{TIC}: Training-Inference Consistency.}
    \vspace{-6pt}

    \resizebox{0.7\linewidth}{!}{
        \begin{tabular}{ccc|cc|cc|c}
            \hlineB{2}
            \multicolumn{3}{c|}{\textbf{Strategy}}  & \multicolumn{2}{c|}{\textbf{Q-Real}} & \multicolumn{2}{c|}{\textbf{AGIQA-3K}} & \multicolumn{1}{c}{\textbf{ImageReward}} \\ \cline{1-8}
            
            \textit{N-CoT} & \textit{D-CoT} & \textit{TIC} & \textit{SRCC} & \textit{PLCC} & \textit{SRCC} & \textit{PLCC} & \textit{Acc}\\ \hline
            
            - & \checkmark & \checkmark & \underline{0.700} & 0.694 & 0.570 & 0.598 &  0.558 \\ 
            \checkmark & - & \checkmark & 0.696 & \underline{0.700} & \underline{0.576} & \underline{0.601} & \underline{0.560} \\ 
            \checkmark & \checkmark & - & 0.692 & 0.686 & 0.550 & 0.574 & 0.557 \\ 
            
            \hline 
            
            \checkmark & \checkmark & \checkmark & \textbf{0.710} & \textbf{0.720} & \textbf{0.608} & \textbf{0.611} & \textbf{0.607} \\
            \hlineB{2}
        \end{tabular}
    }
    \vspace{-12pt}
    \label{tab:quality_assessment_ablation}
\end{table}

\subsection{Ablation Study}
\noindent \textit{\textbf{1)For Q-Real Bench Evaluation.}} We conduct a ablation study to assess the contribution of Stage 1 training on the model's performance in the ImageQA task. The result is presented in Table~\ref{tab:ImageQA_ablation}.  It is clear that performing Stage 1 training first better prepares the model to handle the more complex training task in Stage 2.

\noindent \textit{\textbf{2)For Quality Assessment Evaluation.}} We conduct an ablation study to validate the effectiveness of our prompt design in Stage 3. Specifically, we evaluate the contributions of including naturalness issues, distortion issues, and training-inference consistency in the prompt design. The results are presented in Table~\ref{tab:quality_assessment_ablation}, which show that including both naturalness and distortion issues in the prompt improves quality prediction, with distortion issues having a stronger correlation with quality and training-inference consistency has the most significant impact—models without this strategy perform the worst.

\section{Conclusion}
We present \textbf{Q-Real}, a fine-grained quality assessment dataset of 10K AI-generated images, focusing on naturalness and distortion. Moving beyond traditional score-based metrics, Q-Real integrates entity localization, objective judgments, and subjective analyses. Based on this, we introduce \textbf{Q-Real Bench} for accurate model evaluation about the fine-grained assessment performance and design a dedicated \textbf{training pipeline} that effectively unifies fine-grained assessment with score prediction. Our experimental results demonstrate the effectiveness of both the dataset and our proposed method. We hope our work inspires future advancements in MLLM-based quality assessment for AI-generated content.


%
%
\bibliographystyle{splncs04}
\bibliography{main}
\end{document}